\documentclass[11pt]{article}

\usepackage[final]{acl}

\usepackage{times}
\usepackage{latexsym}
\usepackage{amsmath}
\usepackage{amsmath}
\usepackage{amssymb}
\usepackage{algorithm}
\usepackage{algorithmic}
\usepackage[T1]{fontenc}
\usepackage{algorithm}
\usepackage{algorithmic}  
\usepackage{amsmath}
\usepackage[table]{xcolor}
\usepackage{booktabs}    
\usepackage{multirow}    
\usepackage{array}       

\usepackage{rotating}

\usepackage{rotating}   
\usepackage{tabularx}   
\usepackage{booktabs}
\usepackage{xcolor}

\usepackage{latexsym}
\usepackage{amssymb}
\usepackage{amsmath}
\usepackage{amsthm}
\usepackage{booktabs}
\usepackage{enumitem}
\usepackage{graphicx}
\usepackage{color}

\usepackage{multirow} 
\usepackage{adjustbox} 
\usepackage{xspace}

\usepackage{algorithm}
\usepackage{xcolor}
\definecolor{phasecolor}{RGB}{0, 0, 150} 
\definecolor{commcolor}{RGB}{100, 100, 100} 
\usepackage[utf8]{inputenc}

\usepackage{microtype}

\usepackage{inconsolata}

\usepackage{graphicx}

\title{Parameter Importance is Not Static: Evolving Parameter Isolation for Supervised Fine-Tuning}

\author{
  \textbf{Zekai Lin\textsuperscript{1,$\star$}},
  \textbf{Chao Xue\textsuperscript{5,$\star$}},
  \textbf{Di Liang\textsuperscript{1,2,$\dagger$}},
  \textbf{Xingsheng Han\textsuperscript{1}},
  \textbf{Peiyang Liu\textsuperscript{3}},
  \textbf{Xianjie Wu\textsuperscript{1}},
\\
  \textbf{Lei Jiang\textsuperscript{1}},
  \textbf{Yu Lu\textsuperscript{1}},
  \textbf{Haibo Shi\textsuperscript{1,2}},
  \textbf{Shuang Liang\textsuperscript{4}},
  \textbf{Minlong Peng\textsuperscript{1,$\dagger$}}
\\
\\
\textsuperscript{1} Tencent Hunyuan
    \textsuperscript{2} Tencent Yuanbao
    \textsuperscript{3} Peking University\\
    \textsuperscript{4} UESTC, China
    \textsuperscript{5} University of New South Wales\\
\{rcjgdds,xuechao8071\}@gmail.com; \{liangd17,mlpeng16\}@fudan.edu.cn
}

\begin{document}
\maketitle

\footnotetext[1]{$\star$ Equal Contribution.$\dagger$ Corresponding Author.}
\footnotetext[2]{This work was completed by Zekai Lin and Chao Xue under Di Liang’s supervision.}

\begin{abstract}
Supervised Fine-Tuning (SFT) of large language models often suffers from task interference and catastrophic forgetting. 
Recent approaches alleviate this issue by isolating task-critical parameters during training. 
However, these methods represent a static solution to a dynamic problem, assuming that parameter importance remains fixed once identified.
In this work, we empirically demonstrate that parameter importance exhibits temporal drift over the course of training. To address this, we propose Evolving Parameter Isolation (EPI), a fine-tuning framework that adapts isolation decisions based on online estimates of parameter importance.
Instead of freezing a fixed subset of parameters, EPI periodically updates isolation masks using gradient-based signals, enabling the model to protect emerging task-critical parameters while releasing outdated ones to recover plasticity.
Experiments on diverse multi-task benchmarks demonstrate that EPI consistently reduces interference and forgetting compared to static isolation and standard fine-tuning, while improving overall generalization.
Our analysis highlights the necessity of synchronizing isolation mechanisms with the evolving dynamics of learning diverse abilities.

\end{abstract}

\begin{figure}[t]
  \centering
  \includegraphics[width=1.0\linewidth]{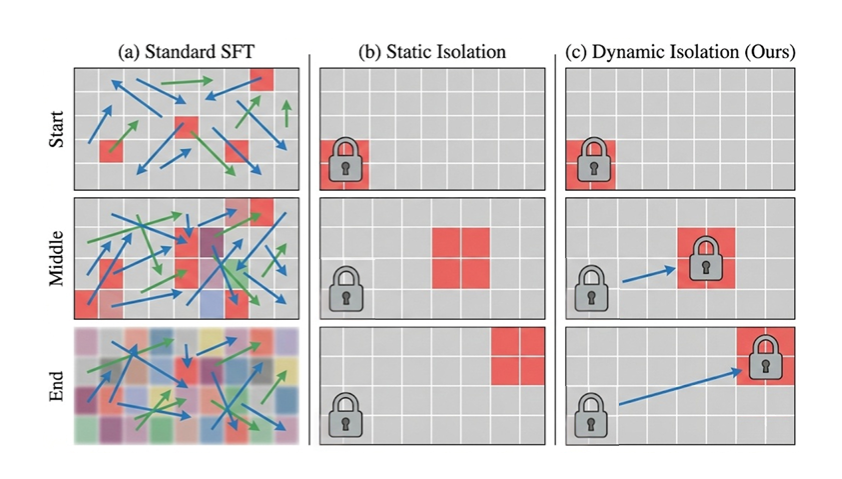} 
  \caption{\textbf{Conceptual illustration of parameter isolation strategies.} 
(a) Standard SFT: parameters are updated globally, leading to interference. Arrows denote task-specific gradient directions.
(b) Static Isolation: The pre-computed mask fails to cover critical parameters (red) as their importance distribution shifts during training.
(c) Dynamic Isolation (Ours): The protection mask is dynamically updated to follow the trajectory of parameter importance.}
  \label{fig:comparison}
\end{figure}

\section{Introduction}

Supervised Fine-Tuning (SFT) has become the dominant paradigm for activating the specialized capabilities of Pre-trained Large Language Models (LLMs), aligning them with diverse applications ranging from complex logical reasoning to open-ended creative generation. While this paradigm has proven effective in single-task scenarios, adapting a single backbone to a heterogeneous task stream presents a fundamental conflict. 
Since all tasks share the same high-dimensional parameter space, the aggressive updates required to learn new capabilities often conflict with the representations of previously acquired knowledge. This parameter-level contention inevitably triggers the ``seesaw phenomenon'' \cite{yu2020gradient} and catastrophic forgetting \cite{mccloskey1989catastrophic, kirkpatrick2017overcoming}, where improvements in one domain (e.g., coding) inadvertently degrade performance in others (e.g., mathematics or dialogue).

To resolve this dilemma, prior research has explored parameter management strategies, building on the premise that different tasks rely on overlapping but distinct subsets of parameter space.
This observation has motivated the development of Parameter-Efficient Fine-Tuning (PEFT) methods. Techniques such as Adapter modules \cite{houlsby2019parameter} and Low-Rank Adaptation (LoRA) \cite{hu2022lora} introduce distinct, modular subspaces to compartmentalize task-specific knowledge. 
More recently, a more direct approach known as Parameter Isolation has emerged as a promising direction. Methods in this category \cite{wang2025not, sorrenti2023selective} explicitly identify task-critical parameters within the original backbone and lock them to prevent destructive interference. 
Despite their empirical success, these isolation approaches rely on a key but largely unexamined assumption: they presume that the importance of parameters is a static property that, once identified at initialization, remains fixed throughout training.

We argue that this static assumption is misaligned with the highly dynamic nature of SFT.
Fine-tuning is inherently an unstable process. As training progresses, the model's internal representations evolve, optimization trajectories shift, and the functional roles of specific parameters change significantly.
We refer to this phenomenon as Parameter Importance Drift.
For instance, parameters responsible for learning basic output formats in the early stages may become redundant later, while those governing complex reasoning logic may only become critical in later training stages.
As visually illustrated in Figure~\ref{fig:comparison}, the set of task-critical parameters may drift over training; we verify this quantitatively later through mask-dynamics analyses.
Consequently, static isolation mechanisms inevitably fall out of sync with the learning trajectory. This misalignment leads to two detrimental consequences: on one hand, the model wastes capacity by protecting obsolete parameters that no longer require safeguarding; on the other hand, it fails to lock emerging critical regions, leaving them vulnerable to being overwritten by subsequent updates.

To bridge this gap, we propose Evolving Parameter Isolation (EPI), a dynamic fine-tuning framework designed to synchronize protection mechanisms with the model's evolving optimization trajectory.
Unlike prior works that rely on pre-computed, static masks, EPI introduces an online importance estimation mechanism that continuously monitors gradient-based signals to track parameter sensitivity and update the isolation mask in real time.
Because instantaneous gradients are noisy and gradient scales differ across layers, EPI combines temporal smoothing (EMA) with layer-wise normalization before selecting the protected subset.
Using this evolving feedback, EPI periodically updates its isolation decisions, effectively creating a ``moving shield'' that adaptively locks currently critical parameters while releasing redundant ones for flexible adaptation.
This mechanism resolves the conflict between Stability (retaining established knowledge) and Plasticity (absorbing new capabilities) by keeping the isolation strategy aligned with the learning trajectory.

Our contributions are summarized as follows:
(1) We identify and quantify the temporal drift in parameter importance during the SFT process, providing empirical evidence that challenges the static assumption in parameter isolation research. 
(2) We propose EPI, a parameter-efficient framework that leverages online gradient statistics to dynamically update isolation masks without introducing extra trainable parameters. 
(3) Extensive experiments across diverse benchmarks covering reasoning, coding, and dialogue demonstrate that EPI consistently outperforms both standard SFT and static isolation baselines, yielding a more stable and generalizable adaptation across diverse tasks.

\begin{figure*}[t!]
  \centering
  \includegraphics[width=1.0\linewidth]{teaser_11.png} 
  \caption{\textbf{Overview of the Evolving Parameter Isolation (EPI) framework.} The model trains on heterogeneous data (Left) while the EPI Engine (Middle) continuously estimates parameter sensitivity using exponential moving averages of squared gradients. These signals drive the Evolutionary Locking mechanism (Right), which dynamically updates isolation masks from $T_{start}$ to $T_{end}$ to protect shifting task-critical parameters.}
  \label{fig:epi_framework}
\end{figure*}

\section{Methodology}
\label{sec:method}

In this section, we present Evolving Parameter Isolation (EPI) (illustrated in Figure~\ref{fig:epi_framework}), a framework explicitly designed to address the dynamic nature of parameter importance during the SFT of LLMs.
We first provide a theoretical analysis of gradient interference via the lens of loss landscape geometry.
Then, we detail the EPI framework, covering online importance estimation, adaptive layer-wise normalization, and the dynamic masking mechanism.

\subsection{Problem Formulation and Gradient Interference}

We consider the SFT of a pre-trained Large Language Model (LLM), parameterized by $\boldsymbol{\theta} \in \mathbb{R}^d$, on a heterogeneous stream of tasks $\mathcal{T}$.
The global optimization objective is to minimize the expected risk across all tasks:
\begin{equation}
\boldsymbol{\theta}^* = \mathop{\arg\min}_{\boldsymbol{\theta}} \sum_{k=1}^{K} \mathbb{E}_{\mathcal{D}_k} \left[ \ell \big( f_{\boldsymbol{\theta}}(x), y \big) \right].
\label{eq:mtl_objective}
\end{equation}

However, simply minimizing this aggregate loss is insufficient due to parameter contention. In this multi-task setting, let $\boldsymbol{g}_i$ and $\boldsymbol{g}_j$ denote the gradients derived from two distinct tasks $\mathcal{T}_i$ and $\mathcal{T}_j$. 
We formally define Task Interference based on the cosine similarity between their update directions:
\begin{equation}
\mathcal{C}(\boldsymbol{g}_i, \boldsymbol{g}_j) = \frac{\boldsymbol{g}_i \cdot \boldsymbol{g}_j}{\|\boldsymbol{g}_i\| \|\boldsymbol{g}_j\|}.
\end{equation}
When $\mathcal{C}(\boldsymbol{g}_i, \boldsymbol{g}_j) < 0$, the gradients are conflicting, leading to the ``seesaw phenomenon'' where reducing the loss for $\mathcal{T}_i$ increases the loss for $\mathcal{T}_j$.
Our goal is to identify a dynamic parameter subspace where such conflicts are minimized.

\subsection{Theoretical Motivation: Minimizing Forgetting}
\label{sec:static_limitation}

To prevent catastrophic forgetting, we must ensure that the loss on previously learned capabilities $\mathcal{L}_{\text{old}}$ does not increase significantly.
Assuming the model has converged to a local optimum $\boldsymbol{\theta}^*$ for previous tasks, we can approximate the loss increase caused by a parameter shift $\Delta \boldsymbol{\theta}$ using a second-order Taylor expansion \cite{kirkpatrick2017overcoming}:
\begin{align} 
\mathcal{L}_{\text{old}}(\boldsymbol{\theta}^* + \Delta \boldsymbol{\theta}) &\approx \mathcal{L}_{\text{old}}(\boldsymbol{\theta}^*) + \Delta \boldsymbol{\theta}^\top \nabla \mathcal{L}_{\text{old}}(\boldsymbol{\theta}^*) \nonumber \\
&\quad + \frac{1}{2} \Delta \boldsymbol{\theta}^\top \mathbf{H} \Delta \boldsymbol{\theta},
\label{eq:taylor}
\end{align}
where $\mathbf{H}$ is the Hessian matrix. Given the assumption of local optimality for previous tasks $\nabla \mathcal{L}_{\text{old}}(\boldsymbol{\theta}^*) \approx \mathbf{0}$, minimizing forgetting is equivalent to minimizing the quadratic term $\frac{1}{2} \Delta \boldsymbol{\theta}^\top \mathbf{H} \Delta \boldsymbol{\theta}$.
This implies that parameters with high curvature (large Hessian values) are critical for preserving knowledge.

Based on this principle, traditional parameter isolation methods identify critical regions at initialization and fix the protection mask $\boldsymbol{m}_t$ throughout training (i.e., $\boldsymbol{m}_t \equiv \boldsymbol{m}_0$). However, we argue that parameter importance is not invariant but drifts over time. To quantify this effect, we define the Temporal Drift as the Hamming distance between ideal masks at different steps:
\begin{equation}
d_H(\boldsymbol{m}^*_{t_1}, \boldsymbol{m}^*_{t_2}) = \sum_{j=1}^d \mathbb{I}\left(m_{t_1}^{(j)} \neq m_{t_2}^{(j)}\right).
\end{equation}
Our analysis (see Appendix) shows that $d_H$ is significant, suggesting that static masks inevitably fall out of sync with the learning trajectory and necessitating a dynamic strategy.

\subsection{Evolving Parameter Isolation (EPI)}

To bridge the gap between static protection and dynamic learning, EPI introduces a time-varying isolation mechanism.

\subsubsection{Online Importance Estimation via Fisher Approximation}
Computing the exact Hessian $\mathbf{H}$ involves high computational and memory costs ($O(d^2)$) for LLMs.
To circumvent this bottleneck, we leverage the theoretical insight that the Fisher Information Matrix (FIM) asymptotically approaches the Hessian near local minima \cite{pascanu2013revisiting, martens2020new}.
To ensure scalability, we adopt the Empirical Fisher approximation, restricting our estimation to the diagonal elements using squared gradients.

Let $\boldsymbol{g}_t = \nabla_{\boldsymbol{\theta}} \mathcal{L}(\boldsymbol{\theta}_t)$. We maintain a running sensitivity vector $\boldsymbol{S}_t \in \mathbb{R}^d$ via Exponential Moving Average (EMA):
\begin{equation}
\boldsymbol{S}_t = \beta \boldsymbol{S}_{t-1} + (1 - \beta) (\boldsymbol{g}_t \odot \boldsymbol{g}_t),
\label{eq:importance_ema}
\end{equation}
where $\beta \in [0, 1)$ is a smoothing coefficient.

\begin{figure}[t]
    \centering
    \includegraphics[width=1\linewidth]{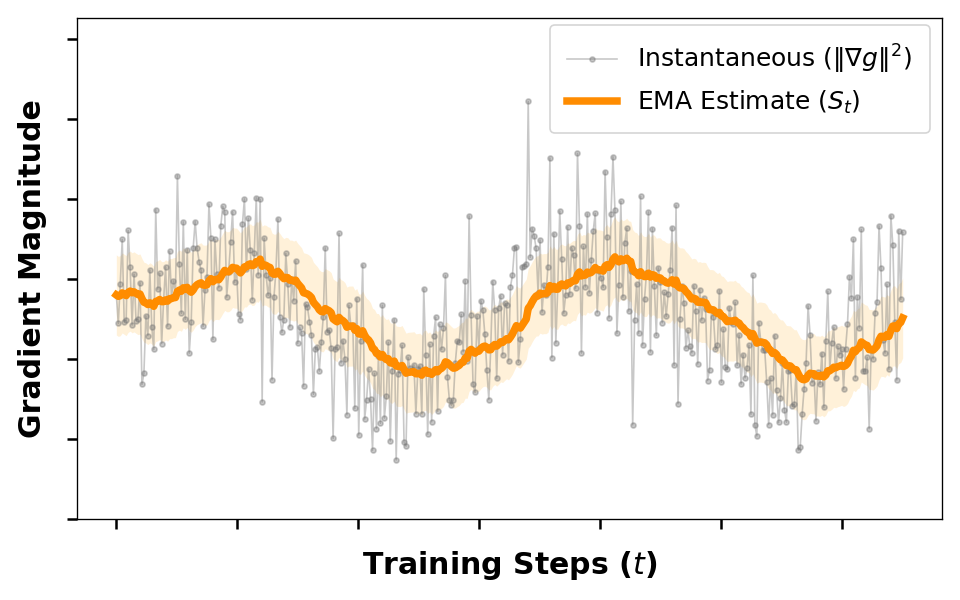} 
    \caption{\textbf{Conceptual illustration of temporal smoothing.} The Instantaneous Gradient (Gray) exhibits high variance. In contrast, the EMA Estimate (Orange) effectively filters high-frequency noise, revealing the underlying importance trend ($\boldsymbol{S}_t$).}
    \label{fig:ema_noise}
\end{figure}

As illustrated in Figure~\ref{fig:ema_noise}, the instantaneous squared gradients ($\boldsymbol{g}_t \odot \boldsymbol{g}_t$) exhibit high variance due to stochastic mini-batch sampling. The EMA mechanism acts as a low-pass filter, suppressing high-frequency noise to reveal the stable underlying importance trend. This formulation effectively approximates the local curvature:
\begin{equation}
S_t^{(j)} \approx \mathbb{E}_{t' \le t} \left[ \left( \frac{\partial \mathcal{L}}{\partial \theta_j} \right)^2 \right] \propto \mathbf{H}_{jj}.
\end{equation}

\subsubsection{Adaptive Layer-wise Normalization}

\begin{figure}[t]
  \centering
  \includegraphics[width=1\linewidth]{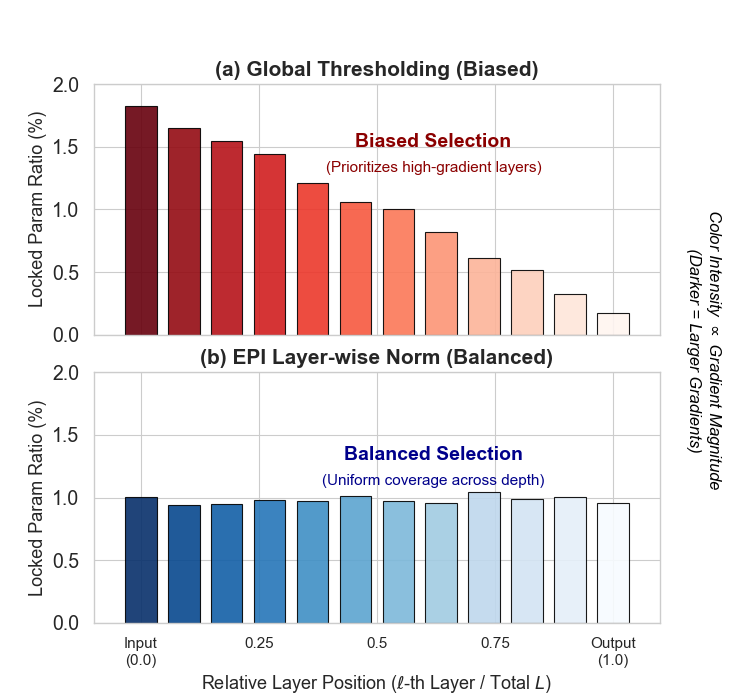}
  \caption{\textbf{Conceptual illustration of necessity of Layer-wise Normalization.} (a) Without normalization, global ranking causes severe bias, concentrating protection only on layers with high gradient norms (Red bars). (b) Our Adaptive Layer-wise Normalization rectifies this by projecting scores to a local scale, resulting in a uniform distribution of critical parameters (Blue bars) regardless of the layer's position.}
  \label{fig:layer_norm}
\end{figure}

Crucially, simply ranking $\boldsymbol{S}_t$ globally is flawed. In deep Transformer architectures, gradient magnitudes vary significantly across layers due to backpropagation dynamics (i.e., the gradient scale mismatch).
Consequently, a global thresholding strategy would disproportionately bias the mask towards specific layers with naturally larger gradients, leaving other critical layers unprotected (as shown in Figure~\ref{fig:layer_norm}).

To ensure unbiased parameter selection, we apply Adaptive Min-Max Normalization.
Let $\Omega_l$ denote the set of parameters in the $l$-th layer. We first compute layer-wise statistics:
\begin{equation}
\mu_{t,l} = \min_{k \in \Omega_l} S_t^{(k)}, \quad \nu_{t,l} = \max_{k \in \Omega_l} S_t^{(k)}.
\end{equation}
The normalized score $\hat{I}_t^{(j)}$ for parameter $j \in \Omega_l$ is then computed as:
\begin{equation}
\hat{I}_t^{(j)} = \frac{S_t^{(j)} - \mu_{t,l}}{\nu_{t,l} - \mu_{t,l} + \epsilon}.
\end{equation}
This projects importance scores into a unified $[0, 1]$ probability space, preserving the relative importance ranking within each layer while normalizing scale differences.

\subsection{Dynamic Mask Generation and Evolution}

Based on $\hat{\boldsymbol{I}}_t$, EPI periodically updates the isolation mask. We define an update interval $H$. At step $t$ (where $t \bmod H = 0$), the binary mask $\boldsymbol{m}_t$ is generated via Top-$p\%$ selection:
\begin{equation}
m_t^{(j)} = \mathbb{I}\left( \hat{I}_t^{(j)} \ge \rho_t(p) \right),
\end{equation}
where $\rho_t(p)$ is the dynamic threshold corresponding to the $(1-p)$-quantile.

\begin{figure}[t]
    \centering
    \includegraphics[width=1\linewidth]{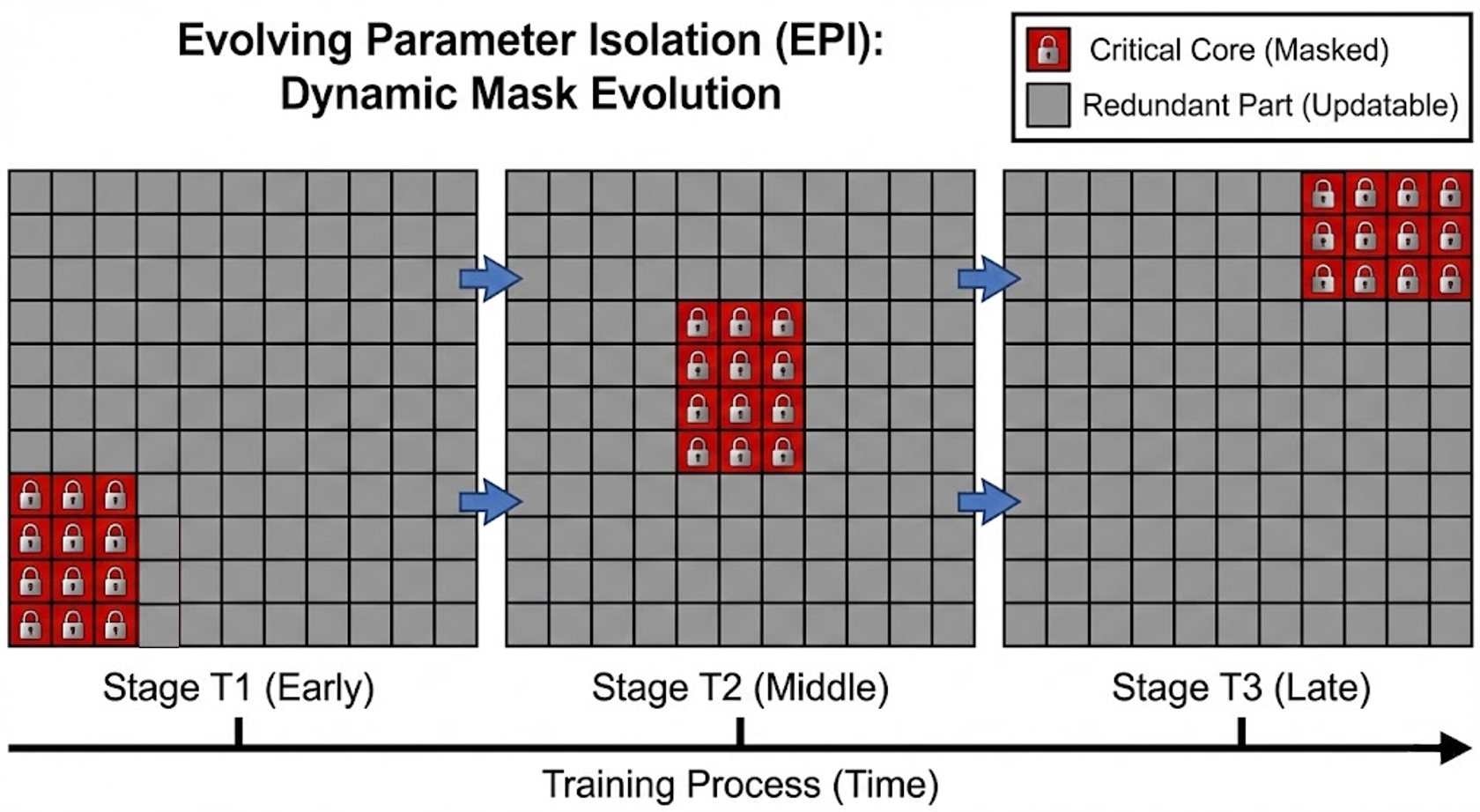} 
    \caption{\textbf{Dynamic Mask Evolution.} Instead of a fixed constraint, the protection mask evolves periodically ($T_1 \rightarrow T_3$). Critical parameters (Red) are locked to prevent forgetting, while redundant ones (Grey) are released. This creates a moving shield that adapts to the temporal drift of parameter importance.}
    \label{fig:mask_evolution}
\end{figure}

Crucially, this mechanism enables the Evolution of Protection. We formally define the parameter state transitions as:
\begin{align}
\mathcal{S}_{\text{lock}}^{(t)} &= \{ j \mid m_t^{(j)} = 1 \land m_{t-1}^{(j)} = 0 \}, \\
\mathcal{S}_{\text{free}}^{(t)} &= \{ j \mid m_t^{(j)} = 0 \land m_{t-1}^{(j)} = 1 \}.
\end{align}
Here, $\mathcal{S}_{\text{lock}}$ represents emerging critical parameters (visualized as the Red Set in Figure~\ref{fig:mask_evolution}), while $\mathcal{S}_{\text{free}}$ denotes parameters that became redundant and are released for plasticity.

\subsection{Optimization with Dynamic Masking}

To enforce isolation under AdamW, we apply the mask to the final parameter update (so protected parameters also receive no decoupled weight decay). 
At step $t$, we first compute the AdamW update:
\begin{equation}
\Delta \boldsymbol{\theta}_t \leftarrow \text{AdamWStep}(\boldsymbol{\theta}_t,\boldsymbol{g}_t),
\end{equation}
and then update parameters with the isolation mask:
\begin{equation}
\boldsymbol{\theta}_{t+1} = \boldsymbol{\theta}_t + (\mathbf{1}-\boldsymbol{m}_t)\odot \Delta \boldsymbol{\theta}_t,
\label{eq:masked_update}
\end{equation}
where $\mathbf{1}$ is the all-ones vector and $\boldsymbol{m}_t \in \{0,1\}^d$.
This blocks updates on protected parameters and mitigates forgetting (Eq.~\ref{eq:taylor}) while preserving adaptation capacity.

\definecolor{rank1}{rgb}{1.0, 0.85, 0.7} 
\definecolor{rank2}{rgb}{1.0, 0.92, 0.8} 
\definecolor{rank3}{rgb}{1.0, 0.97, 0.92}

\newcommand{\cOne}{\cellcolor{rank1}} 
\newcommand{\cTwo}{\cellcolor{rank2}} 
\newcommand{\cThree}{\cellcolor{rank3}}

\begin{table*}[t]
\centering
\small
\setlength{\tabcolsep}{0.35em}
\renewcommand{\arraystretch}{1.3} 

\begin{tabular}{@{}llcccccc@{}}
\toprule
\textbf{Base Model} & \textbf{Method} & \textbf{GSM8K} & \textbf{CodeAlpaca} & \textbf{LogiQA} & \textbf{Alpaca} & \textbf{UltraChat} & \textbf{Avg. Norm.} \\
\midrule

\multirow{5}{*}{LLaMA-3-8B} 
 & Full SFT & 55.6 & 28.3 & 60.8 & 7.8 & 8.1 & 7.32 \\
 & Multi-Stage (Random) & 56.9 & 27.9 & 61.5 & \cThree 8.0 & \cThree 8.2 & 7.44 \\
 & Multi-Stage (Heuristic) & \cThree 57.4 & \cThree 28.7 & \cThree 62.1 & 7.7 & 8.0 & \cThree 7.48 \\
 & Static Isolation ($p=1\%$) & \cTwo 59.2 & \cTwo 29.4 & \cTwo 63.0 & \cTwo 8.1 & \cTwo 8.3 & \cTwo 7.68 \\
 & \textbf{EPI (Ours)} & \cOne \textbf{61.8} & \cOne \textbf{31.2} & \cOne \textbf{65.4} & \cOne \textbf{8.4} & \cOne \textbf{8.6} & \cOne \textbf{7.98} \\
 
\midrule[0.5pt]

\multirow{5}{*}{Mistral-7B}
 & Full SFT & 53.1 & \cThree 26.7 & 58.9 & 7.5 & 7.9 & 7.05 \\
 & Multi-Stage (Random) & 54.4 & 26.3 & 59.6 & \cThree 7.7 & \cThree 8.0 & 7.17 \\
 & Multi-Stage (Heuristic) & \cThree 55.0 & \cTwo 27.1 & \cThree 60.2 & 7.4 & 7.8 & \cThree 7.21 \\
 & Static Isolation ($p=1\%$) & \cTwo 56.8 & 27.9 & \cTwo 61.4 & \cTwo 7.8 & \cTwo 8.1 & \cTwo 7.44 \\
 & \textbf{EPI (Ours)} & \cOne \textbf{59.3} & \cOne \textbf{29.5} & \cOne \textbf{63.1} & \cOne \textbf{8.2} & \cOne \textbf{8.4} & \cOne \textbf{7.74} \\
 
\midrule[0.5pt]

\multirow{5}{*}{Qwen2-7B}
 & Full SFT & 57.2 & 29.6 & 62.4 & 8.0 & 8.3 & 7.55 \\
 & Multi-Stage (Random) & 58.5 & 29.2 & 63.0 & \cThree 8.2 & \cThree 8.4 & 7.67 \\
 & Multi-Stage (Heuristic) & \cThree 59.1 & \cThree 30.0 & \cThree 63.7 & 7.9 & 8.2 & \cThree 7.71 \\
 & Static Isolation ($p=1\%$) & \cTwo 61.0 & \cTwo 30.8 & \cTwo 64.9 & \cTwo 8.3 & \cTwo 8.6 & \cTwo 7.93 \\
 & \textbf{EPI (Ours)} & \cOne \textbf{63.7} & \cOne \textbf{32.6} & \cOne \textbf{67.2} & \cOne \textbf{8.7} & \cOne \textbf{8.9} & \cOne \textbf{8.23} \\

\midrule[0.5pt]

\multirow{5}{*}{Gemma-2-9B}
 & Full SFT & 58.5 & 30.1 & 64.0 & 8.3 & 8.5 & 7.78 \\
 & Multi-Stage (Random) & 59.8 & 29.8 & 64.6 & \cThree 8.5 & \cThree 8.7 & 7.90 \\
 & Multi-Stage (Heuristic) & \cThree 60.4 & \cThree 30.6 & \cThree 65.2 & 8.2 & 8.4 & \cThree 7.97 \\
 & Static Isolation ($p=1\%$) & \cTwo 62.2 & \cTwo 31.5 & \cTwo 66.4 & \cTwo 8.6 & \cTwo 8.8 & \cTwo 8.21 \\
 & \textbf{EPI (Ours)} & \cOne \textbf{65.0} & \cOne \textbf{33.2} & \cOne \textbf{69.1} & \cOne \textbf{8.9} & \cOne \textbf{9.1} & \cOne \textbf{8.57} \\

\bottomrule
\end{tabular}
\caption{\textbf{Main experimental results on heterogeneous SFT benchmarks.}
EPI consistently outperforms standard SFT, multi-stage training, and static parameter isolation across all base models.
Dynamic isolation yields the largest gains on reasoning-intensive tasks (GSM8K, LogiQA) and improves overall generalization as measured by Avg. Norm. Score. Background colors denote the top-3 methods (Darker Orange indicates higher performance).} 
\label{tab:main_results}
\vspace{-1.0em}
\end{table*}

\section{Experimental Setup}
\label{sec:experiments}

We evaluate EPI in a sequential multi-stage supervised fine-tuning setting, where heterogeneous tasks are learned stage by stage rather than jointly mixed within each mini-batch.
We conduct extensive experiments to assess the effectiveness of EPI under this challenging scenario.
Our evaluation is designed to answer the following questions:
(i) whether EPI consistently outperforms standard supervised fine-tuning (SFT) and static parameter isolation baselines under heterogeneous and conflicting task distributions;
(ii) whether dynamically evolving isolation masks are more effective than fixed masks in mitigating task interference and catastrophic forgetting;
(iii) how sensitive EPI is to key hyperparameters such as the isolation ratio $p$ and update interval $H$;
and (iv) whether EPI maintains strong generalization across both structured reasoning and open-ended instruction-following tasks.

\subsection{Datasets}

We evaluate EPI on a diverse collection of publicly available benchmarks covering mathematical reasoning, logical reasoning, code generation, and open-ended instruction following.
This heterogeneous task mixture is intentionally designed to induce parameter contention and expose interference effects.

\noindent\textbf{Mathematical Reasoning.}\quad
We use GSM8K \cite{cobbe2021training}, a widely adopted benchmark for multi-step arithmetic reasoning.
Performance is measured using exact-match accuracy.

\noindent\textbf{Logical Reasoning.}\quad
We evaluate logical consistency using LogiQA \cite{liu2020logiqa}, which requires multi-hop deductive reasoning over short passages.
Accuracy is used as the evaluation metric.

\noindent\textbf{Code Generation.}\quad
For code synthesis, we adopt CodeAlpaca \cite{chaudhary2023code}, following prior work \cite{wang2025not}.
We evaluate generation quality using CodeBLEU \cite{ren2020codebleu}.

\noindent\textbf{Instruction Following and Dialogue.}\quad
We use Alpaca \cite{taori2023stanford} and UltraChat \cite{ding2023enhancing} to assess general instruction-following and conversational abilities.
Following established practice \cite{zheng2023judging}, responses are evaluated using GPT-4-based scoring on a 1–10 scale.

Each dataset is evaluated using its standard metric.
To facilitate a unified comparison across heterogeneous tasks, we additionally report a macro-average score (Avg. Norm. Score) by normalizing all task-specific metrics to a common 0–10 scale.
\subsection{Baselines}
We compare EPI against the following strong SFT baselines, largely following the experimental protocol of \citet{wang2025not} to ensure fairness and comparability.

\noindent\textbf{Full Multi-task SFT.}\quad
The model is fine-tuned on a uniform mixture of all datasets, updating all parameters jointly without task grouping or isolation.
This represents the standard instruction tuning paradigm and serves as a lower bound to quantify the severity of task interference in the shared parameter space.

\noindent\textbf{Multi-Stage SFT (Random Grouping).}\quad
Tasks are randomly partitioned into separate stages. The model is fine-tuned sequentially on each stage, updating all parameters throughout training.
This setting simulates a basic sequential learning scenario, allowing us to evaluate the model's vulnerability to catastrophic forgetting when task order is randomized.

\noindent \textbf{Multi-Stage SFT (Heuristic Grouping).}\quad
Tasks are grouped based on high-level semantic similarity (e.g., reasoning tasks vs.\ open-ended dialogue) and fine-tuned sequentially, with all parameters updated in each stage.
This baseline tests whether manual curriculum design can mitigate interference more effectively than random ordering, or if parameter isolation is necessary.

\noindent \textbf{Static Parameter Isolation.}\quad
We implement static parameter isolation following \citet{sorrenti2023selective, wang2025not}, where a fixed subset of task-critical parameters is identified during an initial probing phase and frozen throughout subsequent training.
Crucially, this method relies on a pre-computed mask derived from initial statistics, assuming that parameter importance remains invariant to the optimization trajectory—an assumption directly challenged by our EPI framework.

\subsection{Implementation Details}

We evaluate all methods using recent, strong open-weight LLMs as backbone models, updating earlier baselines to reflect current practice.
Specifically, we use LLaMA-3-8B \cite{dubey2024llama}, Qwen2-7B \cite{yang2024qwen2}, Gemma-2-9B \cite{team2024gemma} and Mistral-7B \cite{jiang2023mistral7b} as base models.
All models are fine-tuned using the AdamW optimizer \cite{loshchilov2017decoupled} with a learning rate of $1 \times 10^{-5}$, batch size of 64, and a cosine learning rate schedule with 3\% warm-up steps.
Unless otherwise stated, all SFT procedures are conducted for three epochs.
For static isolation baselines, task-critical parameters are identified via probe fine-tuning runs of one epoch per task, following \citet{wang2025not}.
The isolation ratio is set to $p=1\%$.
For EPI, importance statistics are updated online using exponential moving averages of squared gradients (Eq.~\ref{eq:importance_ema}).
Isolation masks are refreshed every $H=500$ optimization steps, with a default isolation ratio of $p=1\%$.
All experiments are conducted on eight NVIDIA A100 GPUs (80GB).
To ensure reproducibility and robustness, we conduct each experiment over three independent runs with different random seeds and report the average performance.

\subsection{Main Results}
\label{sec:main_results}

Table~\ref{tab:main_results} summarizes the performance of EPI compared to standard SFT and various isolation baselines across four state-of-the-art backbone models. The results support three key observations:

\noindent\textbf{EPI consistently establishes a new state-of-the-art across heterogeneous benchmarks.} \quad
As shown in Table~\ref{tab:main_results}, EPI achieves the highest Average Normalized Score across all four base models, consistently outperforming the standard Full SFT baseline by a significant margin (e.g., +0.79 average score improvement on Gemma-2-9B). Notably, simple Multi-Stage SFT strategies (both Random and Heuristic) often fail to yield consistent improvements over Full SFT, and in some cases (e.g., Mistral-7B on CodeAlpaca), sequential training even degrades performance due to catastrophic forgetting. This confirms that merely reordering data or strictly separating training stages is insufficient to resolve parameter contention in the shared parameter space.

\noindent\textbf{Dynamic evolution outperforms static protection, especially on complex reasoning.} \quad
The most critical comparison lies between EPI and Static Isolation, as both methods aim to protect task-critical parameters. Our results demonstrate that EPI consistently surpasses Static Isolation across all tasks. Crucially, the performance gap is most pronounced in reasoning-intensive tasks. For instance, on GSM8K, EPI outperforms Static Isolation by +2.6\% on LLaMA-3-8B and +2.8\% on Gemma-2-9B. We attribute this to the fact that reasoning capabilities often require the formation of complex functional circuits that evolve during training. Static masks, fixed at initialization, fail to capture these emerging dependencies. By contrast, EPI’s ability to dynamically lock emerging critical parameters allows the model to secure new reasoning pathways as they are learned.

\noindent\textbf{EPI effectively mitigates the Stability-Plasticity dilemma.} \quad
A common failure mode in static parameter isolation is over-protection, where locking too many parameters hinders the learning of new open-ended tasks (Plasticity). However, EPI achieves superior performance not only on rigid reasoning tasks but also on creative generation tasks (Alpaca, UltraChat). For example, on Qwen2-7B, EPI improves UltraChat performance to 8.9, surpassing both Full SFT (8.3) and Static Isolation (8.6). This indicates that EPI successfully releases outdated parameters (as detailed in Sec.~\ref{sec:method}), recovering sufficient capacity to adapt to open-ended instructions without overwriting established knowledge. This balance is reflected in the Average Normalized Score, where EPI demonstrates the most robust generalization capability across diverse task categories.

\section{Ablation Study}
\label{sec:ablation}

To disentangle the contributions of individual components within the EPI framework, we conduct a comprehensive ablation study using LLaMA-3-8B and Gemma-2-9B.
We systematically investigate the following key aspects:
(1) the necessity of dynamic evolution and layer-wise normalization;
(2) the superiority of our selection mechanism over alternative strategies;
(3) the impact of temporal gradient smoothing via EMA;
(4) the sensitivity to critical hyperparameters, specifically the isolation ratio $p$ and the mask-update interval $H$;
and (5) the underlying temporal and structural dynamics of the evolving masks.
Unless otherwise stated, we report the Avg. Norm. Score to facilitate a unified comparison across heterogeneous tasks.

\subsection{Component Effectiveness Analysis}
\label{sec:ablation_components}

We first validate the core premise of EPI: namely, that parameter isolation must be dynamic and normalized.
Table~\ref{tab:mask_ablation} compares the full EPI method against a static baseline and a variant without layer-wise normalization.

\begin{table}[h!]
\centering
\small
\renewcommand{\arraystretch}{1.4}
\begin{tabular*}{\linewidth}{@{\extracolsep{\fill}}lcc@{}}
\toprule
\textbf{Mask Strategy} & \textbf{LLaMA-3} & \textbf{Gemma-2} \\
\midrule
Static Mask ($p=1\%$) & 7.68 & 8.21 \\
Static Mask + Layer Norm & 7.82 & 8.36 \\
EPI (w/o Layer Norm) & 7.83 & 8.35 \\
\textbf{EPI (Full)} & \textbf{7.98} & \textbf{8.57} \\
\bottomrule
\end{tabular*}
\caption{\textbf{Ablation of mask strategy.} Both layer-wise normalization and dynamic evolution contribute meaningfully, and their combination yields the strongest performance.}
\label{tab:mask_ablation}
\end{table}

\noindent\textbf{Complementary roles of dynamic evolution and layer-wise normalization.}\quad
Table~\ref{tab:mask_ablation} shows that both components contribute meaningfully and are complementary.
Adding layer-wise normalization to a static mask already yields clear gains on both backbones (LLaMA-3: $7.68 \rightarrow 7.82$; Gemma-2: $8.21 \rightarrow 8.36$), confirming that normalization alleviates cross-layer gradient-scale bias even without mask evolution.
Conversely, enabling dynamic evolution without layer normalization also improves performance over the static baseline (LLaMA-3: $7.68 \rightarrow 7.83$; Gemma-2: $8.21 \rightarrow 8.35$), showing that periodic re-selection is itself effective in tracking temporal drift.

\noindent\textbf{Why the full design works best.}\quad
The strongest results are achieved only when both components are used together, reaching 7.98 on LLaMA-3 and 8.57 on Gemma-2.
This indicates that the gains from EPI do not come from a single design choice alone.
Instead, layer-wise normalization improves the quality and fairness of importance scores across layers, while dynamic evolution improves their temporal adaptivity throughout training.
Their combination yields the most effective protection mechanism under heterogeneous SFT.

\subsection{Alternative Selection Strategies}
\label{sec:ablation_alt_selection}

To better justify our normalization and selection design, we compare EPI against two simpler alternatives.
The first is Per-Layer Fixed Budget, which selects the top-$p\%$ parameters independently within each layer.
The second is Global Raw Scores, which ranks all parameters globally using unnormalized importance scores.
Table~\ref{tab:alt_selection} summarizes the results on LLaMA-3-8B.

\begin{table}[h!]
\centering
\small
\renewcommand{\arraystretch}{1.4}
\resizebox{\columnwidth}{!}{%
\begin{tabular}{lc}
\toprule
\textbf{Strategy} & \textbf{Avg.\ Norm.\ Score} \\
\midrule
Per-Layer Fixed Budget & 7.82 \\
Global Raw Scores      & 7.71 \\
\textbf{EPI (Ours)}    & \textbf{7.98} \\
\bottomrule
\end{tabular}%
}
\caption{\textbf{Comparison of alternative allocation strategies on LLaMA-3-8B.} Both per-layer fixed quotas and global raw ranking underperform our layer-wise normalization plus global top-$p\%$ selection.}
\label{tab:alt_selection}
\end{table}

Both simpler alternatives are inferior to the full EPI design.
Per-layer fixed budgets impose a rigid quota regardless of each layer's actual sensitivity and plasticity, which can lead to misallocation of the protection budget.
By contrast, global raw ranking is biased toward layers with naturally larger gradient magnitudes and therefore over-protects high-scale layers.
EPI performs best because it first removes cross-layer scale bias through layer-wise normalization and then allocates the protection budget globally, allowing protection to concentrate where it is most needed.

\subsection{Impact of Gradient Smoothing (EMA)}

\begin{table}[h!]
\centering
\small
\renewcommand{\arraystretch}{1.5}
\begin{tabular*}{\linewidth}{@{\extracolsep{\fill}}lcc@{}}
\toprule
\textbf{Smoothing Factor} & \textbf{LLaMA-3} & \textbf{Gemma-2} \\
\midrule
No EMA (Instantaneous) & 7.81 & 8.39 \\
EMA ($\beta=0.9$) & 7.90 & 8.50 \\
\textbf{EMA ($\beta=0.99$)} & \textbf{7.98} & \textbf{8.57} \\
\bottomrule
\end{tabular*}
\caption{\textbf{Effect of Temporal Accumulation.} A high smoothing factor ($\beta=0.99$) is essential to filter stochastic noise and reveal stable importance trends.}
\label{tab:ema_ablation}
\end{table}

The results indicate that raw instantaneous gradients ($\beta=0$) are insufficient for reliable parameter isolation due to high variance.
Notably, performance peaks at $\beta=0.99$, which empirically validates that strong temporal smoothing is essential to filter out transient fluctuations and reveal consistent importance trends.
This confirms that the isolation mechanism benefits from capturing the long-term optimization trajectory rather than reacting to stochastic gradient noise.

\subsection{Sensitivity to Isolation Ratio ($p$)}

\begin{figure}[h!]
\centering
\includegraphics[width=1\linewidth]{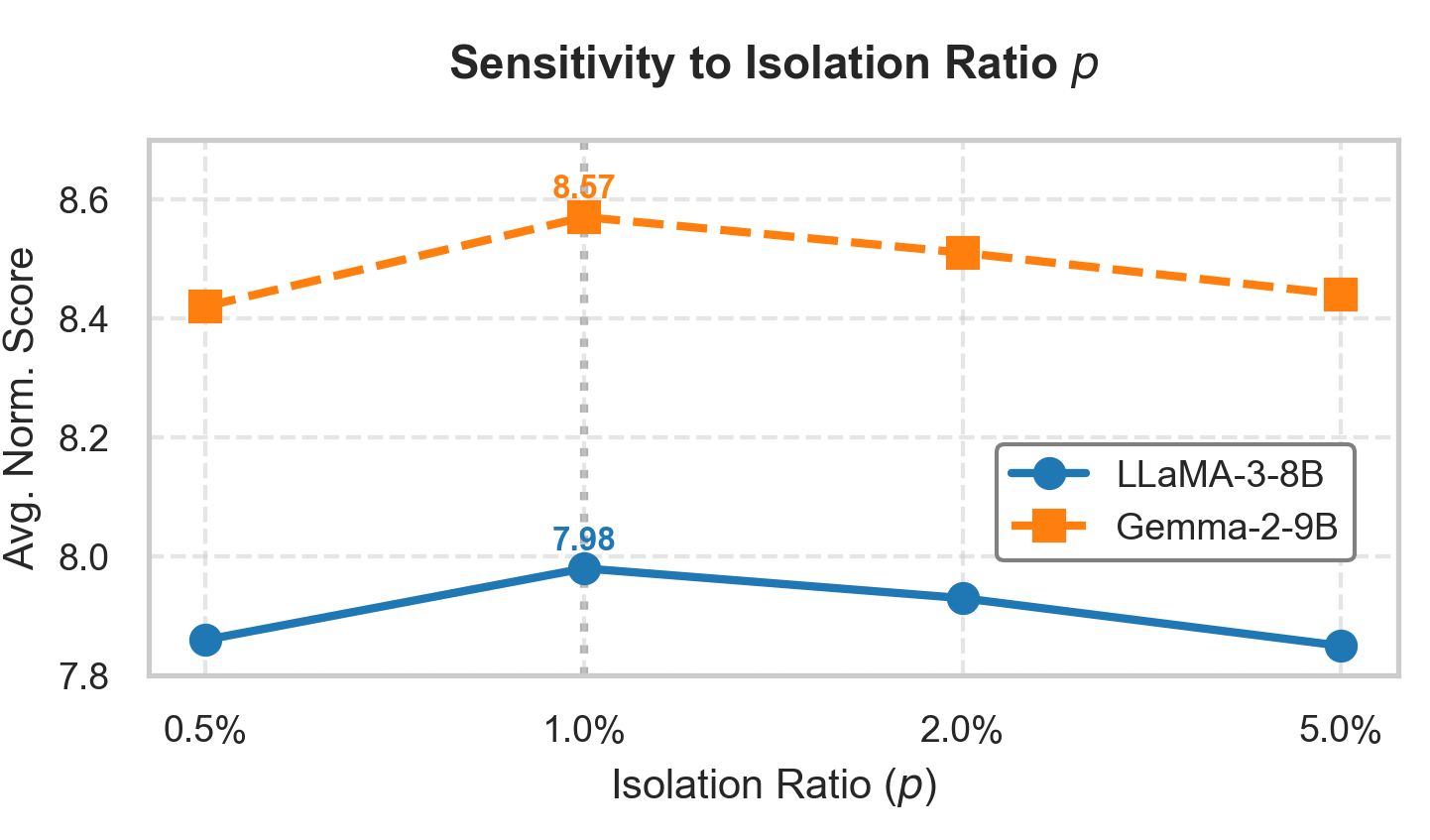}
\caption{\textbf{Sensitivity Analysis to Isolation Ratio ($p$).} The plot illustrates the average normalized score across different isolation ratios for LLaMA-3 and Gemma-2. Both models exhibit a distinct peak performance at $p=1.0\%$, representing an optimal trade-off between retaining prior knowledge (Stability) and adapting to new tasks (Plasticity).}
\label{fig:core_sensitivity}
\end{figure}

Finally, we investigate the crucial trade-off between Stability and Plasticity by varying the core parameter isolation ratio $p$. The results are visualized in Figure~\ref{fig:core_sensitivity}.

As illustrated in Figure~\ref{fig:core_sensitivity}, we observe a clear inverted U-shaped trend consistent across both model architectures. Performance peaks significantly at the critical value of $p=1.0\%$, where LLaMA-3 achieves a score of 7.98 and Gemma-2 reaches 8.57.

Deviating from this optimal operating point leads to diminished returns, highlighting the delicate balance required. Lower ratios ($p=0.5\%$) result in a shield too sparse to ensure stability against forgetting, whereas higher ratios ($p \ge 2.0\%$) introduce excessive rigidity that limits the plasticity needed for new task adaptation. This finding, where the optimal balance is struck at a mere $p=1\%$, provides evidence that task-critical knowledge in LLMs is highly sparse, residing in only a very small fraction of important parameters.

\subsection{Sensitivity to Mask-Update Interval}
\label{sec:ablation_H}

Although the importance statistic $S_t$ is updated continuously, the binary protection mask is refreshed only every $H$ steps. 
This design trades off responsiveness against computational overhead and optimization stability. 
Very frequent updates can introduce threshold thrashing for parameters near the top-$p\%$ boundary, while overly infrequent updates lag behind temporal drift.

\begin{table}[h!]
\centering
\small
\renewcommand{\arraystretch}{1.2}
\resizebox{\columnwidth}{!}{%
\begin{tabular}{lc}
\toprule
\textbf{Update Interval ($H$)} & \textbf{Avg.\ Norm.\ Score} \\
\midrule
100  & 7.92 \\
\textbf{500}  & \textbf{7.98} \\
1000 & 7.95 \\
2000 & 7.89 \\
\bottomrule
\end{tabular}%
}
\caption{\textbf{Sensitivity to the mask-update interval $H$ on LLaMA-3-8B.} $H=500$ provides the best trade-off between responsiveness and stability.}
\label{tab:H_main}
\end{table}

As shown in Table~\ref{tab:H_main}, performance is robust across a reasonable range of update intervals, with a slight peak at $H=500$. 
This supports our design choice of periodic rather than per-step mask updates: updating too frequently introduces instability, while updating too slowly reduces adaptability to drifting parameter importance.

\subsection{Quantifying Mask Dynamics}
\label{sec:mask_dynamics}

To verify the existence of Parameter Importance Drift, we track the evolution of isolation masks throughout the training process.

\noindent \textbf{Temporal Divergence.} \quad
We first examine the stability of parameter importance over time. Concretely, on LLaMA-3-8B, the Jaccard overlap between the initial mask and the final mask drops from 78.3\% at 25\% training progress to 44.7\% at the end of training, indicating that more than half of the initially protected parameters are replaced over time (see Appendix~\ref{sec:appendix_mask_dynamics}). This divergence indicates that many parameters deemed critical at initialization become redundant as the model adapts. Consequently, static isolation approaches, which enforce a rigid lock based on initial statistics, inevitably waste capacity on outdated parameters while failing to secure newly emerging functional circuits.

\noindent \textbf{Structural and Task Adaptation.} \quad
Beyond global drift, the mask evolution exhibits clear structural patterns. We observe that upper layers (closer to the output) show significantly faster evolution rates compared to lower layers (see Appendix \ref{sec:appendix_mask_dynamics}).
This aligns with the intuition that upper layers manage complex, task-specific reasoning which requires high plasticity, while lower layers handle stable linguistic features.
Furthermore, the mask shows distinct reconfiguration patterns immediately following task transitions, confirming that EPI is actively responsive to data distribution shifts rather than drifting randomly.

\section{Conclusion}
\label{sec:conclusion}

In this paper, we investigated the dynamic nature of parameter importance during supervised fine-tuning (SFT) of large language models and identified the limitations of static isolation methods.  
We proposed \textbf{Evolving Parameter Isolation (EPI)}, a novel framework that continuously tracks parameter sensitivity via gradient-based online estimation, applies adaptive layer-wise normalization, and evolves protection masks to dynamically align with the learning trajectory.  
Extensive experiments across multiple LLM backbones and heterogeneous tasks including reasoning, code generation, and instruction following demonstrated that EPI consistently outperforms standard SFT and static isolation baselines.  
Our analyses show that EPI not only improves overall task performance but also significantly reduces catastrophic forgetting and gradient interference (the seesaw effect), highlighting the importance of synchronizing isolation mechanisms with the evolving optimization dynamics of SFT.  
Our findings suggest that future fine-tuning paradigms must move beyond rigid static constraints and embrace dynamic, adaptive mechanisms to fully unlock the potential of LLMs in heterogeneous multi-task scenarios. 

\newpage
\section{Limitations}
\label{sec:limitations}

Despite its strong empirical performance, EPI has several limitations that should be considered in future work.  

First, while the online estimation of parameter importance using EMA of squared gradients is computationally efficient, it introduces additional memory overhead proportional to the model size. For extremely large models (e.g., 30B+ parameters), this could limit scalability without further optimization or sparsity-aware techniques.  

Second, EPI currently relies on heuristic hyperparameters such as the core percentage $p$ and mask update interval $H$. Although we show robustness across a range of values, these parameters may require careful tuning for different models or task distributions. An adaptive mechanism to auto-tune $p$ based on loss landscapes would be a valuable extension. 

Third, our experiments focus primarily on English-centric datasets and structured reasoning, code, and dialogue tasks. The effectiveness of EPI on low-resource languages, highly multimodal data, or continual adaptation with streaming data remains to be validated.  

Finally, while dynamic isolation mitigates gradient interference, it does not explicitly model task similarity or hierarchy. Future extensions could incorporate task-aware importance propagation or knowledge distillation techniques to further enhance cross-task generalization.

\bibliography{custom}

\appendix
\section{Related Work}

\subsection{Challenges in Heterogeneous SFT}
Supervised Fine-Tuning (SFT) has become a core paradigm for adapting pre-trained large language models (LLMs) to downstream instructions and tasks \cite{touvron2023llama, brown2020language, radford2019language, minaee2024large, vaswani2017attention, howard2018universal}. Building on this foundation, recent instruction-tuning approaches have substantially improved the generalization ability of LLMs across diverse tasks \cite{wei2021finetuned, sanh2021multitask, chung2024scaling, ouyang2022training, wang2023self,gao2026decorl}. Meanwhile, the rapid expansion of model scale, data scale, and task diversity has further broadened the scope of capabilities that LLMs are expected to support \cite{lu2023makes, luo2023wizardcoder,liu2024resolving,liu2025structural}.

A major challenge, however, is that modern SFT must accommodate an increasingly heterogeneous capability space within a single shared parameter space. Compared with earlier NLP settings centered on semantic matching and natural language inference \cite{wang2022dabert, xue2023dual, song2022improving, li2024local, liang2019asynchronous,li2024comateformer}, contemporary LLMs are expected to handle substantially more diverse tasks, including structured table reasoning \cite{wu2025breaking, wu2025tablebench,wu2026mmtablebench}, knowledge-intensive dialogue generation \cite{wu2025unleashing}, temporal and multi-hop reasoning \cite{xue2024question, liu2023time, liu2023local, fei2022cqg}, optimization-oriented problem solving \cite{wang2026rethinking}, and stable long-range dependency modeling \cite{dai2025hope}. These capabilities differ markedly in input structure, output format, and optimization objective, making them difficult to reconcile through uniform parameter updates.

As a result, SFT over heterogeneous task streams often suffers from gradient conflict and update competition. Updates that improve one capability may degrade another, leading to the ``seesaw phenomenon'' \cite{yu2020gradient}, shifts in ability composition under different fine-tuning data mixtures \cite{dong2024abilities}, and eventually catastrophic forgetting \cite{mccloskey1989catastrophic, kirkpatrick2017overcoming}.

\subsection{Traditional Mitigation Paradigms}
To address these conflicts, prior literature has predominantly explored two directions: global optimization within shared spaces and physical separation via modular architectures.

\noindent\textbf{Global Optimization in Shared Parameter Spaces.}\quad
Strategies in this category attempt to reconcile conflicts by adjusting the optimization trajectory of the entire model, treating parameters as a unified entity. 
Data-centric approaches utilize heuristic curriculum learning \cite{ouyang2022training, wei2021finetuned,wu2025progressive} or dynamic scheduling \cite{aribandi2021ext5} to balance task exposure. To mitigate forgetting in sequential settings, replay mechanisms like DMT \cite{dong2024abilities} and experience replay \cite{rolnick2019experience} introduce historical data into current training stages.
Complementarily, gradient-based methods constrain update directions. Regularization techniques such as Elastic Weight Consolidation (EWC) \cite{kirkpatrick2017overcoming,li2017learning} penalize changes to important weights, while projection-based gradient constraints (or gradient surgery) such as GEM \cite{lopez2017gem} and PCGrad \cite{yu2020gradient} modify gradients to reduce conflicts. Separately, gradient normalization methods such as GradNorm \cite{chen2018gradnorm} adaptively reweight tasks by matching gradient magnitudes. However, finding a single update direction for conflicting objectives remains geometrically infeasible, failing to resolve destructive interference.

\noindent\textbf{Architectural Modularization.}\quad
To avoid contention in shared spaces, a distinct paradigm physically decouples task representations by expanding the model architecture. This includes Mixture-of-Experts (MoE) \cite{fedus2022switch, riquelme2021scaling}, Adapter Fusion \cite{pfeiffer2021adapterfusion}, and split-architecture designs \cite{stickland2019bert}. While effective at reducing interference through separation, these modular approaches introduce significant structural complexity and memory overhead. Furthermore, manually assigning static modules often fails to scale efficiently to hundreds of fine-grained capabilities, potentially leading to knowledge fragmentation \cite{cai2025survey}.

\begin{table*}[t!]
\centering
\small
\setlength{\tabcolsep}{0pt}
\begin{tabular*}{\textwidth}{@{\extracolsep{\fill}} llcccccc @{}}
\toprule
\multirow{2}{*}{\textbf{Base Model}} & \multirow{2}{*}{\textbf{Method}} & \multicolumn{5}{c}{\textbf{Forgetting Ratio (FR, \%) $\downarrow$}} & \multirow{2}{*}{\textbf{Avg. FR}} \\
\cmidrule(lr){3-7}
 &  & \textbf{GSM8K} & \textbf{Code} & \textbf{LogiQA} & \textbf{Alpaca} & \textbf{UChat} &  \\
\midrule
\multirow{5}{*}{LLaMA-3-8B} 
 & Full SFT (Multi-task) & 12.6 & 10.4 & 9.8 & 7.5 & 6.9 & 9.44 \\
 & Multi-Stage (Random) & 9.7 & 8.9 & 8.1 & 6.8 & 6.2 & 7.94 \\
 & Multi-Stage (Heuristic) & 8.9 & 8.2 & 7.5 & 6.5 & 6.0 & 7.42 \\
 & Static Isolation ($p=1\%$) & 6.1 & 5.8 & 5.2 & 4.5 & 4.1 & 5.14 \\
 & \textbf{EPI (Ours, $p=1\%$)} & \textbf{5.3} & \textbf{5.1} & \textbf{4.8} & \textbf{4.2} & \textbf{3.9} & \textbf{4.66} \\
\midrule
\multirow{5}{*}{Gemma-2-9B} 
 & Full SFT (Multi-task) & 11.2 & 9.7 & 9.0 & 7.3 & 7.1 & 8.66 \\
 & Multi-Stage (Random) & 8.5 & 7.9 & 7.3 & 6.5 & 6.0 & 7.04 \\
 & Multi-Stage (Heuristic) & 7.9 & 7.2 & 6.8 & 6.2 & 5.7 & 6.36 \\
 & Static Isolation ($p=1\%$) & 5.6 & 5.3 & 4.9 & 4.1 & 3.8 & 4.74 \\
 & \textbf{EPI (Ours, $p=1\%$)} & \textbf{4.8} & \textbf{4.5} & \textbf{4.2} & \textbf{3.8} & \textbf{3.4} & \textbf{4.14} \\
\midrule
\multirow{5}{*}{Qwen2-7B} 
 & Full SFT (Multi-task) & 13.0 & 11.5 & 10.2 & 8.0 & 7.5 & 10.04 \\
 & Multi-Stage (Random) & 10.1 & 9.1 & 8.5 & 6.9 & 6.4 & 8.20 \\
 & Multi-Stage (Heuristic) & 9.4 & 8.5 & 8.0 & 6.7 & 6.1 & 7.74 \\
 & Static Isolation ($p=1\%$) & 6.8 & 6.3 & 5.8 & 4.9 & 4.5 & 5.66 \\
 & \textbf{EPI (Ours, $p=1\%$)} & \textbf{5.7} & \textbf{5.0} & \textbf{4.7} & \textbf{4.1} & \textbf{3.8} & \textbf{4.66} \\
\bottomrule
\end{tabular*}
\caption{Forgetting analysis across tasks and base models. Lower FR (\%) indicates better retention of previously learned knowledge. EPI consistently reduces forgetting compared to standard SFT and static isolation baselines.}
\label{tab:forgetting_aligned}
\end{table*}

\subsection{Parameter-Level Isolation Techniques}
Recent research has pivoted towards a third paradigm: Parameter-Level Isolation. Grounded in the theory of parameter heterogeneity \cite{neyshabur2014search, frankle2018lottery, mu2025comprehensive}, these methods aim to construct logical subspaces within the native backbone, offering a balance between the efficiency of shared models and the stability of modular designs.

\noindent\textbf{Implicit Subspaces via Additive Modules.}\quad
Parameter-Efficient Fine-Tuning (PEFT) methods, exemplified by Adapters \cite{houlsby2019parameter,li2021prefix} and LoRA \cite{hu2022lora}, implicitly achieve isolation by freezing the backbone and optimizing task-specific low-rank matrices. While this effectively separates task knowledge, it fundamentally relies on additive parameters. This design choice leaves the vast majority of the pre-trained dense knowledge dormant, potentially limiting the model's capacity to leverage its full reasoning power during adaptation.

\noindent\textbf{Explicit Partitioning and Static Isolation.}\quad
Distinct from additive methods, explicit isolation strategies partition the existing parameters of the pre-trained backbone. Early works in transfer learning explored iterative pruning to specialize weights \cite{mallya2018packnet, parmar2024reuse,zheng2022robust}. More recently, Core Parameter Isolation Fine-Tuning (CPI-FT) \cite{wang2025not} advanced this concept by explicitly identifying and freezing task-critical parameters based on their update magnitudes.
However, a critical limitation persists in these approaches: they rely on a static assumption. Current methods typically fix the isolation mask at initialization, ignoring the temporal dynamics of the optimization process. As training progresses and feature hierarchies reorganize, parameters initially deemed critical may become redundant, while previously unimportant parameters may emerge as essential \cite{liu2026dpi,evci2020rigging}.

\noindent\textbf{Evolving Parameter Isolation (Ours).}\quad
Our proposed EPI framework addresses this gap by transitioning from static to dynamic isolation. Unlike coarse-grained routing \cite{qi2024optimizing} or static freezing mechanisms \cite{wang2025not}, EPI utilizes online gradient signals to continuously update the protection mask. By synchronizing the isolation mechanism with the model's evolving learning trajectory, EPI ensures that emerging capabilities are protected in real-time while maintaining the flexibility of the shared parameter space.

\begin{table*}[t!]
\centering
\small
\setlength{\tabcolsep}{0pt}
\begin{tabular*}{\textwidth}{@{\extracolsep{\fill}} llcccccc @{}}
\toprule
\multirow{2}{*}{\textbf{Base Model}} & \multirow{2}{*}{\textbf{Method}} & \multicolumn{5}{c}{\textbf{Task Gradient Conflict (TGC) $\downarrow$}} & \multirow{2}{*}{\textbf{Avg. TGC}} \\
\cmidrule(lr){3-7}
 &  & \textbf{Code} & \textbf{LogiQA} & \textbf{Alpaca} & \textbf{Code-Logi} & \textbf{Code-Alp} &  \\
\midrule
\multirow{5}{*}{LLaMA-3-8B} 
 & Full SFT (Multi-task) & 0.42 & 0.38 & 0.36 & 0.33 & 0.30 & 0.358 \\
 & Multi-Stage (Random) & 0.31 & 0.28 & 0.26 & 0.24 & 0.22 & 0.262 \\
 & Multi-Stage (Heuristic) & 0.28 & 0.25 & 0.23 & 0.21 & 0.20 & 0.234 \\
 & Static Isolation ($p=1\%$) & 0.18 & 0.15 & 0.13 & 0.12 & 0.11 & 0.138 \\
 & \textbf{EPI (Ours, $p=1\%$)} & \textbf{0.15} & \textbf{0.12} & \textbf{0.10} & \textbf{0.09} & \textbf{0.08} & \textbf{0.108} \\
\midrule
\multirow{5}{*}{Gemma-2-9B} 
 & Full SFT (Multi-task) & 0.39 & 0.36 & 0.33 & 0.30 & 0.28 & 0.332 \\
 & Multi-Stage (Random) & 0.28 & 0.26 & 0.23 & 0.22 & 0.20 & 0.238 \\
 & Multi-Stage (Heuristic) & 0.25 & 0.23 & 0.21 & 0.19 & 0.18 & 0.212 \\
 & Static Isolation ($p=1\%$) & 0.14 & 0.12 & 0.11 & 0.09 & 0.08 & 0.108 \\
 & \textbf{EPI (Ours, $p=1\%$)} & \textbf{0.13} & \textbf{0.11} & \textbf{0.09} & \textbf{0.08} & \textbf{0.07} & \textbf{0.096} \\
\midrule
\multirow{5}{*}{Qwen2-7B} 
 & Full SFT (Multi-task) & 0.44 & 0.40 & 0.37 & 0.35 & 0.32 & 0.376 \\
 & Multi-Stage (Random) & 0.33 & 0.30 & 0.27 & 0.25 & 0.22 & 0.274 \\
 & Multi-Stage (Heuristic) & 0.30 & 0.27 & 0.25 & 0.23 & 0.20 & 0.250 \\
 & Static Isolation ($p=1\%$) & 0.17 & 0.15 & 0.13 & 0.11 & 0.10 & 0.132 \\
 & \textbf{EPI (Ours, $p=1\%$)} & \textbf{0.16} & \textbf{0.13} & \textbf{0.11} & \textbf{0.09} & \textbf{0.08} & \textbf{0.114} \\
\bottomrule
\end{tabular*}
\caption{Seesaw effect analysis via Task Gradient Conflict (TGC). Lower TGC signifies reduced task interference. EPI minimizes gradient conflicts more effectively than full multi-task and static baselines.}

\label{tab:seesaw_aligned}
\end{table*}

\section{Forgetting and Seesaw Effect Analysis}
\label{sec:forgetting}

Catastrophic forgetting and task interference constitute fundamental challenges in multi-task supervised fine-tuning (SFT) of large language models (LLMs). When models undergo sequential or simultaneous fine-tuning on heterogeneous tasks, parameter updates intended for a specific task often degrade previously acquired capabilities—a phenomenon manifesting as the ``seesaw effect''. To mitigate these issues, our proposed Evolving Parameter Isolation (EPI) explicitly identifies and safeguards task-critical parameters dynamically, while permitting redundant parameters to adapt to novel tasks.

\subsection{Evaluation Metrics}

We employ two primary metrics to systematically quantify forgetting and task interference:

\paragraph{Forgetting Ratio (FR).} For a given task $\mathcal{T}_i$, FR quantifies the degradation in performance on previously learned tasks following fine-tuning on new tasks:
\begin{equation}
    \text{FR}_i = \frac{\text{Perf}_{\text{initial}}(\mathcal{T}_i) - \text{Perf}_{\text{after new tasks}}(\mathcal{T}_i)}{\text{Perf}_{\text{initial}}(\mathcal{T}_i)} \times 100\%.
\end{equation}

\paragraph{Task Gradient Conflict (TGC).} TGC measures the severity of gradient interference between any task pair $\mathcal{T}_i$ and $\mathcal{T}_j$:
\begin{equation}
    \text{TGC}_{i,j} = \max \big(0, - \cos(\boldsymbol{g}_i, \boldsymbol{g}_j) \big),
\end{equation}
where $\boldsymbol{g}_i$ and $\boldsymbol{g}_j$ denote the gradient vectors of the respective tasks. A higher TGC value indicates stronger interference, reflecting a pronounced seesaw effect.

We conduct sequential fine-tuning experiments starting with GSM8K (mathematical reasoning), followed by CodeAlpaca (code generation), LogiQA (logical reasoning), and finally Alpaca and UltraChat (instruction-following and dialogue). This sequence establishes a controlled environment to assess catastrophic forgetting on early tasks and gradient conflicts across diverse stages. We evaluate three backbones: LLaMA-3-8B, Gemma-2-9B, and Qwen2-7B. Baselines include Full Multi-task SFT, Multi-Stage Random grouping, and Multi-Stage Heuristic grouping. All models are trained under identical optimization and batch configurations to ensure fair comparison.

\subsection{Results on Forgetting and Seesaw Effect}

The experimental results, summarized in Table \ref{tab:forgetting_aligned} and Table \ref{tab:seesaw_aligned}, yield several key observations. 
First, Full Multi-task SFT suffers from substantial forgetting and high gradient conflicts, attributable to indiscriminate parameter updates across heterogeneous tasks. 
Second, while Multi-stage baselines partially mitigate forgetting by isolating training stages, static grouping strategies fail to account for the emergence of new critical parameters, leaving sensitive regions vulnerable to interference.
Third, EPI substantially lowers both FR and TGC across all backbones. This demonstrates that dynamically adapting isolation masks effectively preserves previously learned capabilities while facilitating the acquisition of new tasks.

Notably, Gemma-2-9B exhibits consistently lower FR and TGC compared to LLaMA-3-8B. This suggests that larger, more capable models benefit disproportionately from dynamic isolation, likely due to their higher capacity to represent multiple task-specific subspaces. The sequential evolution of the mask in EPI allows the model to track temporal drift in parameter importance, reducing the unnecessary locking of outdated parameters while ensuring emerging critical regions are protected. This dynamic balancing of stability and plasticity directly addresses the seesaw effect, ensuring robust performance in multi-task settings.

\begin{table*}[t!]
\centering
\small
\renewcommand{\arraystretch}{1.2} 
\begin{tabularx}{\textwidth}{@{}p{0.12\textwidth} p{0.22\textwidth} X X X@{}}
\toprule
\textbf{Task} & \textbf{Input Prompt} & \textbf{Full SFT Output} & \textbf{Static Mask Output} & \textbf{EPI Output (Ours)} \\
\midrule
\textbf{GSM8K} \newline (Reasoning) & A train leaves city A at 9:00 AM... (Time/Speed problem) & 1:30 PM & 1:35 PM & 1:30 PM, with step-by-step reasoning showing distance covered by each train. \\
\cmidrule{1-5}
\textbf{CodeAlpaca} \newline (Python) & Write a function that returns Fibonacci numbers up to n. & Returns first 5 numbers only, ignores n. & Correct function but inefficient loop, $O(n^2)$. & Correct and efficient function using iterative approach, handles all n. \\
\cmidrule{1-5}
\textbf{Alpaca} \newline (Instruction) & Summarize the plot of 'Pride and Prejudice' in 2-3 sentences. & Misses key relationships, inconsistent with timeline. & Basic summary, minor omissions. & Concise and accurate summary including major characters, timeline, and main conflict. \\
\cmidrule{1-5}
\textbf{UltraChat} \newline (Dialogue) & User: How do I cook pasta perfectly? & Overgeneralized advice, misses boiling details. & Correct steps but unordered. & Step-by-step guidance with tips for timing, salt, and texture; natural conversational tone. \\
\bottomrule
\end{tabularx}
\caption{Qualitative comparison of model outputs. EPI demonstrates superior reasoning consistency and completeness compared to baselines, effectively mitigating the trade-off between forgetting and learning.}
\label{tab:case_study_adaptive}
\end{table*}

\section{Case Study}
\label{sec:case_study}

To provide a qualitative understanding of how EPI improves SFT, we examine representative cases across reasoning, code generation, and instruction-following tasks. Table~\ref{tab:case_study_adaptive} compares outputs from Full Multi-task SFT, Static Mask Isolation, and our proposed EPI method. 

As shown in the qualitative examples, Full Multi-task SFT often suffers from partial correctness and inconsistent reasoning, likely due to interference among heterogeneous tasks. Static Mask Isolation reduces catastrophic forgetting by protecting pre-identified critical parameters; however, it occasionally fails to adapt to emerging task-specific needs, resulting in incomplete or inefficient outputs. 

In contrast, EPI dynamically tracks parameter importance. In GSM8K, EPI produces correct step-by-step reasoning rather than just a final answer. In CodeAlpaca, it generates fully functional, efficient code, whereas static isolation produces suboptimal loops. For Instruction-following, EPI outputs are coherent and contextually appropriate. These cases provide intuitive evidence that dynamic parameter isolation directly contributes to improved task performance and generalization by balancing stability and plasticity.

\section{Algorithm and Complexity Analysis}

The complete procedure of Evolving Parameter Isolation (EPI) is summarized in Algorithm~\ref{alg:epi}. 
EPI augments standard SFT with a lightweight online sensitivity tracker and a periodically updated isolation mask. 
At each step, the method estimates parameter importance from gradient magnitudes, updates a binary mask every $H$ steps to protect the most sensitive parameters, and only applies optimization updates to the remaining parameters. 
This design provides a simple mechanism to balance stability (preserving prior knowledge) and plasticity (adapting to new tasks) without introducing additional networks or task-specific heads.

\textbf{Memory Overhead.} 
EPI stores the sensitivity accumulator $\boldsymbol{S}_t$, which has the same dimensionality as the model parameters and therefore incurs a linear memory overhead of $O(d)$, where $d$ is the number of parameters. 
In practice, this is comparable to maintaining one extra optimizer state (e.g., the second-moment estimate in Adam/AdamW). 
Since modern fine-tuning is typically dominated by activation memory (especially under long sequences and large batch sizes), the additional storage for $\boldsymbol{S}_t$ is minor.

\textbf{Computational Efficiency.}
The element-wise EMA update and masking are $O(d)$ operations, incurring negligible overhead relative to the matrix multiplications in forward/backward passes.
EPI supports memory sharding in distributed settings (e.g., DeepSpeed ZeRO), and since the quantile selection is triggered only every $H$ steps, the amortized complexity remains essentially identical to standard AdamW-based SFT.

\textbf{Notes on Hyperparameters.}
The sparsity ratio $p$ and EMA factor $\beta$ regulate protection intensity and estimate smoothness, respectively; specifically, a higher $\beta$ filters stochastic noise from mini-batches.
The interval $H$ balances responsiveness with overhead, where a smaller $H$ allows the model to adapt rapidly to task transitions, while a larger $H$ favors stability in stationary environments.

\begin{table*}[h]
\centering
\small
\setlength{\tabcolsep}{4pt}
\renewcommand{\arraystretch}{1.12}
\begin{tabular*}{\textwidth}{@{\extracolsep{\fill}}llcccccc@{}}
\toprule
\textbf{Base Model} & \textbf{Method} & \textbf{GSM8K} & \textbf{CodeAlpaca} & \textbf{LogiQA} & \textbf{Alpaca} & \textbf{UltraChat} & \textbf{Avg. Norm.} \\
\midrule
\multirow{3}{*}{LLaMA-3-8B}
 & Full SFT & 55.6$\pm$0.4 & 28.3$\pm$0.5 & 60.8$\pm$0.5 & 7.8$\pm$0.2 & 8.1$\pm$0.1 & 7.32$\pm$0.06 \\
 & Static Isolation & 59.2$\pm$0.6 & 29.4$\pm$0.4 & 63.0$\pm$0.4 & 8.1$\pm$0.1 & 8.3$\pm$0.1 & 7.68$\pm$0.04 \\
 & \textbf{EPI (Ours)} & \textbf{61.8$\pm$0.3} & \textbf{31.2$\pm$0.3} & \textbf{65.4$\pm$0.5} & \textbf{8.4$\pm$0.1} & \textbf{8.6$\pm$0.1} & \textbf{7.98$\pm$0.03} \\
\midrule
\multirow{3}{*}{Gemma-2-9B}
 & Full SFT & 58.5$\pm$0.5 & 30.1$\pm$0.6 & 64.0$\pm$0.6 & 8.3$\pm$0.2 & 8.5$\pm$0.2 & 7.78$\pm$0.07 \\
 & Static Isolation & 62.2$\pm$0.4 & 31.5$\pm$0.5 & 66.4$\pm$0.5 & 8.6$\pm$0.1 & 8.8$\pm$0.1 & 8.21$\pm$0.05 \\
 & \textbf{EPI (Ours)} & \textbf{65.0$\pm$0.4} & \textbf{33.2$\pm$0.5} & \textbf{69.1$\pm$0.4} & \textbf{8.9$\pm$0.1} & \textbf{9.1$\pm$0.1} & \textbf{8.57$\pm$0.04} \\
\bottomrule
\end{tabular*}
\caption{\textbf{Main results with standard deviations (3 runs).} EPI consistently outperforms baselines with small standard deviations across tasks, demonstrating that the gains reported in the main text are statistically robust rather than artifacts of a favorable seed.}
\label{tab:main_results_std}
\end{table*}
\section{Additional Experimental Results}
\label{sec:appendix_results}

\subsection{Statistical Robustness}
\label{sec:appendix_std}

To assess robustness, we report mean $\pm$ standard deviation over three independent runs for the main results on LLaMA-3-8B and Gemma-2-9B.

\newcommand{\PHASE}[1]{\STATE \textbf{\textit{\textcolor{phasecolor}{// #1}}}}
\renewcommand{\algorithmiccomment}[1]{\hfill \textcolor{commcolor}{$\triangleright$ \textit{#1}}}

\begin{algorithm}[H]
\caption{Evolving Parameter Isolation (EPI)}
\label{alg:epi}
\begin{algorithmic}[1]
\REQUIRE Pre-trained weights $\boldsymbol{\theta}_0$, Dataset $\mathcal{D}$, Sparsity ratio $p$, EMA factor $\beta$, Interval $H$, Learning rate $\eta_t$.
\ENSURE Optimized parameters $\boldsymbol{\theta}_{T}$.

\STATE \textbf{Initialize:} $\boldsymbol{S}_0 \leftarrow \mathbf{0}$, $\boldsymbol{m}_0 \leftarrow \mathbf{0}$ \COMMENT{Initial mask: all trainable}

\FOR{$t = 1$ \textbf{to} $T_{\text{steps}}$}
    \STATE \textbf{Minibatch Sampling:} $(x, y) \sim \mathcal{D}$.
    \STATE \textbf{Gradient Computation:} $\boldsymbol{g}_t \leftarrow \nabla_{\boldsymbol{\theta}} \mathcal{L}(f_{\boldsymbol{\theta}_{t-1}}(x), y)$.
    
    \vspace{0.3em}
    \PHASE{Phase 1: Online Sensitivity Accumulation}
    \STATE $\boldsymbol{S}_t \leftarrow \beta \boldsymbol{S}_{t-1} + (1 - \beta) (\boldsymbol{g}_t \odot \boldsymbol{g}_t)$
    
    \vspace{0.3em}
    \PHASE{Phase 2: Dynamic Mask Evolution}
    \IF{$t \equiv 0 \pmod H$}
        \STATE $\hat{\boldsymbol{I}} \leftarrow \text{LayerMinMax}(\boldsymbol{S}_t)$
        \COMMENT{Layer-wise Normalization}
        \STATE $\tau \leftarrow \text{Quantile}(\hat{\boldsymbol{I}}, 1-p)$ \COMMENT{Compute dynamic threshold}
        \STATE $\boldsymbol{m}_t \leftarrow \mathbb{I}(\hat{\boldsymbol{I}} \ge \tau)$ \COMMENT{Update binary isolation mask}
    \ELSE
        \STATE $\boldsymbol{m}_t \leftarrow \boldsymbol{m}_{t-1}$ \COMMENT{Retain previous mask}
    \ENDIF
    
    \vspace{0.3em}
    \PHASE{Phase 3: Sparse Optimization}
    \STATE $\Delta\boldsymbol{\theta}_t \leftarrow \text{AdamWStep}(\boldsymbol{\theta}_{t-1}, \boldsymbol{g}_t, \eta_t)$ \COMMENT{Compute AdamW update}
    \STATE $\boldsymbol{\theta}_{t} \leftarrow \boldsymbol{\theta}_{t-1} + (\mathbf{1}-\boldsymbol{m}_t)\odot \Delta\boldsymbol{\theta}_t$ \COMMENT{Mask the update}

\ENDFOR
\end{algorithmic}
\end{algorithm}

The results in Table~\ref{tab:main_results_std} show that the gains of EPI are stable across random seeds. In particular, EPI consistently outperforms the strongest baseline, Static Isolation, on both LLaMA-3-8B and Gemma-2-9B, with relatively small standard deviations across tasks. This suggests that the improvements reported in the main text are robust rather than artifacts of random initialization or data order.

\begin{table*}[t!]
\centering
\small
\setlength{\tabcolsep}{4pt}
\renewcommand{\arraystretch}{1.12}

\begin{tabular*}{\textwidth}{@{\extracolsep{\fill}}lcccccc@{}}
\toprule
\textbf{Isolation Ratio $p$} & \textbf{GSM8K} & \textbf{CodeAlpaca} & \textbf{LogiQA} & \textbf{Alpaca} & \textbf{UltraChat} & \textbf{Avg. Norm.} \\
\midrule
0.5\%  & 58.7 & 28.9 & 62.3 & 8.0 & 8.2 & 7.59 \\
\textbf{1\% }   & \textbf{59.2} & \textbf{29.4} & \textbf{63.0} & \textbf{8.1} & \textbf{8.3} & \textbf{7.68} \\
2\%    & 58.4 & 28.5 & 61.8 & 7.8 & 7.9 & 7.42 \\
5\%    & 56.9 & 27.6 & 60.1 & 7.4 & 7.6 & 7.12 \\
\bottomrule
\end{tabular*}
\caption{\textbf{Static Isolation Ratio Sweep on LLaMA-3-8B}. Static isolation peaks at $p=1\%$ and drops sharply for $p \ge 2\%$, directly supporting over-protection as a failure mode of static masks.}
\label{tab:static_sweep}
\end{table*}

\subsection{Additional Mask Dynamics Analysis}
\label{sec:appendix_mask_dynamics}

To further quantify temporal drift, we analyze the evolution of isolation masks from three complementary perspectives: temporal overlap with the initial mask, layer-wise flip rates, and task-conditioned overlap across datasets. Together, these analyses show that mask evolution is substantial over time, structured across layers, and task-dependent across benchmarks.

\begin{table}[h!]
\centering
\renewcommand{\arraystretch}{1.1}
\resizebox{\columnwidth}{!}{%
\begin{tabular}{lcccc}
\toprule
\textbf{Training Progress} & 25\% & 50\% & 75\% & \textbf{100\% (Final)} \\
\midrule
\textbf{Mask Overlap (\%)} & 78.3 & 62.1 & 51.4 & \textbf{44.7} \\
\bottomrule
\end{tabular}%
}
\caption{\textbf{Temporal drift in isolation masks.} The Jaccard overlap between the initial mask and later masks steadily decreases during training.}
\label{tab:mask_drift_app}
\end{table}

Table~\ref{tab:mask_drift_app} shows direct evidence of temporal drift, with initial mask overlap falling below 50\% by training's end. Although Eq. 4 uses Hamming distance globally, we employ Jaccard similarity to evaluate the sparse active subsets to avoid background zero dominance. This ongoing parameter replacement explains why static masks inevitably misalign with the optimization trajectory.

\begin{table}[h!]
\centering
\small
\resizebox{\columnwidth}{!}{%
\begin{tabular}{lc}
\toprule
\textbf{Layer Group} & \textbf{Average Mask Flip Rate (\%)} \\
\midrule
Layers 1--8   & 12.3 \\
Layers 9--16  & 21.7 \\
Layers 17--24 & 33.5 \\
Layers 25--32 & 41.2 \\
\bottomrule
\end{tabular}%
}
\caption{\textbf{Layer-wise mask evolution.} Upper layers exhibit higher flip rates, indicating stronger adaptation to evolving task demands.}
\label{tab:layer_flip_app}
\end{table}

Table~\ref{tab:layer_flip_app} further shows that mask evolution is not uniform across the network. Upper layers exhibit substantially higher flip rates than lower layers, suggesting that higher-level reasoning and task-specific behaviors require more plasticity, whereas lower layers remain comparatively stable.

\begin{table}[h!]
\centering
\small
\resizebox{\columnwidth}{!}{%
\begin{tabular}{lc}
\toprule
\textbf{Task Pair} & \textbf{Jaccard Similarity (Top-1\%)} \\
\midrule
GSM8K vs.\ CodeAlpaca  & 0.31 \\
GSM8K vs.\ LogiQA      & 0.42 \\
CodeAlpaca vs.\ LogiQA & 0.28 \\
\bottomrule
\end{tabular}%
}
\caption{\textbf{Task-conditioned overlap of important parameters.} Different tasks rely on partially distinct critical parameter subsets.}
\label{tab:task_overlap_app}
\end{table}

Finally, Table~\ref{tab:task_overlap_app} shows that different tasks rely on only partially overlapping sets of important parameters. The relatively low Jaccard similarities confirm that heterogeneous tasks induce different critical regions, further motivating a dynamic isolation mechanism that can adapt as the task distribution changes.

\subsection{Additional Analysis on Static Isolation}
\label{sec:appendix_static}

To complement the $p$-sensitivity analysis of EPI in the main text, we also evaluate static isolation under different protection ratios.

Table~\ref{tab:static_sweep} shows that static isolation also follows an inverted-U trend with respect to the isolation ratio $p$. Performance peaks at $p=1\%$ and degrades sharply at larger ratios, indicating that excessive static locking harms adaptation. Compared with the smoother behavior of EPI in the main text, this result suggests that static masks are more vulnerable to over-protection because they cannot release outdated parameters once training progresses.

\begin{table}[h!]
\centering
\small
\resizebox{\columnwidth}{!}{%
\begin{tabular}{lcccc}
\toprule
\textbf{Method} & $p=0.5\%$ & $p=1\%$ & $p=2\%$ & $p=5\%$ \\
\midrule
Static Isolation & 8.2 & 8.3 & 7.9 & 7.6 \\
EPI             & 8.4 & 8.6 & 8.4 & 8.2 \\
\bottomrule
\end{tabular}%
}
\caption{\textbf{UltraChat performance vs.\ isolation ratio}. EPI maintains higher new-task plasticity at large $p$ by releasing outdated parameters, whereas static isolation exhibits substantially larger degradation.}
\label{tab:plasticity}
\end{table}

The degradation is especially visible on UltraChat, which appears late in the training sequence and therefore serves as a sensitive indicator of remaining plasticity. As shown in Table~\ref{tab:plasticity}, EPI remains substantially more stable at larger isolation ratios, consistent with its release-and-reselect mechanism that preserves a fixed budget while reallocating protection over time.

\subsection{Diagnostic Validation of the Online Importance Criterion}
\label{sec:appendix_diagnostic}

To directly verify that the online importance signal remains informative under temporal drift, we compare it with the true old-task sensitivity estimated through empirical perturbations. 
Concretely, at multiple training stages, we sample a subset of parameters (stratified across layers). We then apply small Gaussian perturbations to these parameters and compute the induced loss change ($\Delta \mathcal{L}_{\text{old}}$) on a fixed, held-out old-task batch. This yields a per-parameter empirical sensitivity signal, which we correlate with our online importance score.

\begin{table}[h!]
\centering
\small
\resizebox{\columnwidth}{!}{%
\begin{tabular}{lcc}
\toprule
\textbf{Training Stage} & \textbf{Pearson $r$} & \textbf{Spearman $\rho$} \\
\midrule
Early  & 0.72 & 0.68 \\
Middle & 0.75 & 0.71 \\
Late   & 0.73 & 0.69 \\
\bottomrule
\end{tabular}%
}
\caption{\textbf{Correlation between online importance and old-task sensitivity (LLaMA-3-8B).} We sample parameters across layers, estimate true old-task sensitivity via small Gaussian perturbations (measured as $\Delta \mathcal{L}_{\text{old}}$ on a held-out batch), and correlate it with our online importance score. Strong correlations persist across all training stages.}
\label{tab:correlation}
\end{table}

As shown in Table~\ref{tab:correlation}, the online importance score remains strongly correlated with old-task sensitivity throughout training. Importantly, the correlation (both Pearson $r$ and Spearman $\rho$) remains stable from early to late stages. This suggests that the EMA-based importance signal successfully tracks meaningful, sensitive coordinates even under substantial parameter drift, supporting our interpretation of EPI as a practical local stability surrogate rather than a static, one-shot estimate.

\end{document}